\documentclass[letterpaper, 10 pt, conference]{ieeeconf}  

\IEEEoverridecommandlockouts                              

\usepackage{graphics} 
\usepackage{epsfig} 
\usepackage{mathptmx} 
\usepackage{times} 
\usepackage{amsmath} 
\usepackage{amssymb}  
\usepackage{subfigure}
\usepackage{floatrow}
\usepackage{float}
\usepackage{booktabs}
\usepackage{array, makecell}

\title{\LARGE \bf
Bridging Scene Understanding and Task Execution with Flexible Simulation Environments
}

\author{Zachary Ravichandran, J. Daniel Griffith, Benjamin Smith, and Costas Frost
\thanks{Z. Ravichandran, J. D. Griffith, and C. Frost are with MIT Lincoln Laboratory, Lexington, MA, USA. B. Smith is with Systems and Technology Research, Woburn, MA, USA, work conducted while at MIT Lincoln Laboratory. E-mail: \{zachary.ravichandran, dan.griffith. constantine.frost\}@ll.mit.edu,  benjamin.smith@stresearch.com}%
\thanks{DISTRIBUTION STATEMENT A. Approved for public release. Distribution is unlimited. This material is based upon work supported by the Under Secretary of Defense for Research and Engineering under Air Force Contract No.  FA8702-15-D-0001.  Any opinions, findings, conclusions or recommendations expressed in this material are those of the author(s) and do not necessarily reflect the views of the Under Secretary of Defense for Research and Engineering}%
}

\begin{document}

\maketitle
\thispagestyle{empty}
\pagestyle{empty}

\begin{abstract}

    Significant progress has been made in scene understanding which seeks to build 3D, metric and object-oriented representations of the world. Concurrently, reinforcement learning has made impressive strides largely enabled by advances in simulation. Comparatively, there has been less focus in simulation for perception algorithms. Simulation is becoming increasingly vital as sophisticated perception approaches such as metric-semantic mapping or 3D dynamic scene graph generation require precise 3D, 2D, and inertial information in an interactive environment. To that end, we present TESSE (Task Execution with Semantic Segmentation Environments), an open source simulator for developing scene understanding and task execution algorithms. TESSE has been used to develop state-of-the-art solutions for metric-semantic mapping and 3D dynamic scene graph generation. Additionally, TESSE served as the platform for the GOSEEK Challenge at the International Conference of Robotics and Automation (ICRA) 2020, an object search competition with an emphasis on reinforcement learning. Code for TESSE is available at https://github.com/MIT-TESSE.

\end{abstract}

\section{INTRODUCTION}

Scene understanding extends traditional Simultaneous Localization and Mapping (SLAM) approaches to provide information including the 3D geometry of a scene, semantic content, and topological representations. This capability produces a rich representation of the environment that may be used for many autonomous robotic tasks including navigation and obstacle avoidance. Several subsystems must work in concert to achieve such a capability. An agent must be able to localize itself using available sensors through methods such as visual-inertial odometry (VIO). The 3D structure of a scene must be estimated. These estimates must then be fused over time to produce a global map. Semantic predictions, such as those from a neural network, may be incorporated to produce a metric-semantic map. Building a robust system presents several challenges as each component is complex and subject to noise. Furthermore, the data needed to develop and evaluate such a system requires fidelity and interactivity that is beyond the scope of many existing datasets. 

In a parallel line of research, the success of deep Reinforcement Learning (RL) on games such as Go and Atari has convinced researchers to attempt to solve real-world tasks~\cite{Silver_2016, mnih2015humanlevel}. Going beyond video games, RL has been applied to several physical tasks including quadrupedal robot locomotion~\cite{Haarnoja2018SoftAA}, unmanned aircraft control~\cite{cad2rl}, and autonomous driving~\cite{DBLP:journals/corr/abs-1804-09364, learning_to_drive_in_a_day}. Since the collection of real world data, especially in the interactive manner required for training, is expensive, time consuming, and in some cases dangerous, simulation has been a critical driver for recent success in RL~\cite{savva2019habitat, Mahmood2018BenchmarkingRL}.
Compared to RL, scene understanding has relied less on simulation in lieu of static datasets~\cite{Burri25012016}. While real world data complexity and noise is difficult to emulate in simulation, real-world datasets often lack the fine-grain detail required to develop and evaluate complex perception systems. 

To solve some of these challenges, we present TESSE (Task Execution with Semantic Segmentation Environments): a simulation framework for developing perception capabilities in conjunction with high level task execution. TESSE provides a rich set of environments and sensors. It has been used to develop state of the art methods for metric-semantic mapping and 3D dynamic scene graph generation. Along with supporting perception development, TESSE is used for RL applications. For example, it was used for the GOSEEK Challenge at ICRA 2020, a competition where participants developed agents to find objects placed within indoor environments. 

In Section \ref{sec:related_work} we discuss related simulators. Section \ref{sec:simulator} provides a more detailed overview of the TESSE simulator. Finally, Section \ref{sec:applications} demonstrates four applications of TESSE spanning from perception to task execution.

\begin{figure*}[t!]
    \centering
    \centerline{\includegraphics[width=0.9\textwidth, height=0.4\textheight]{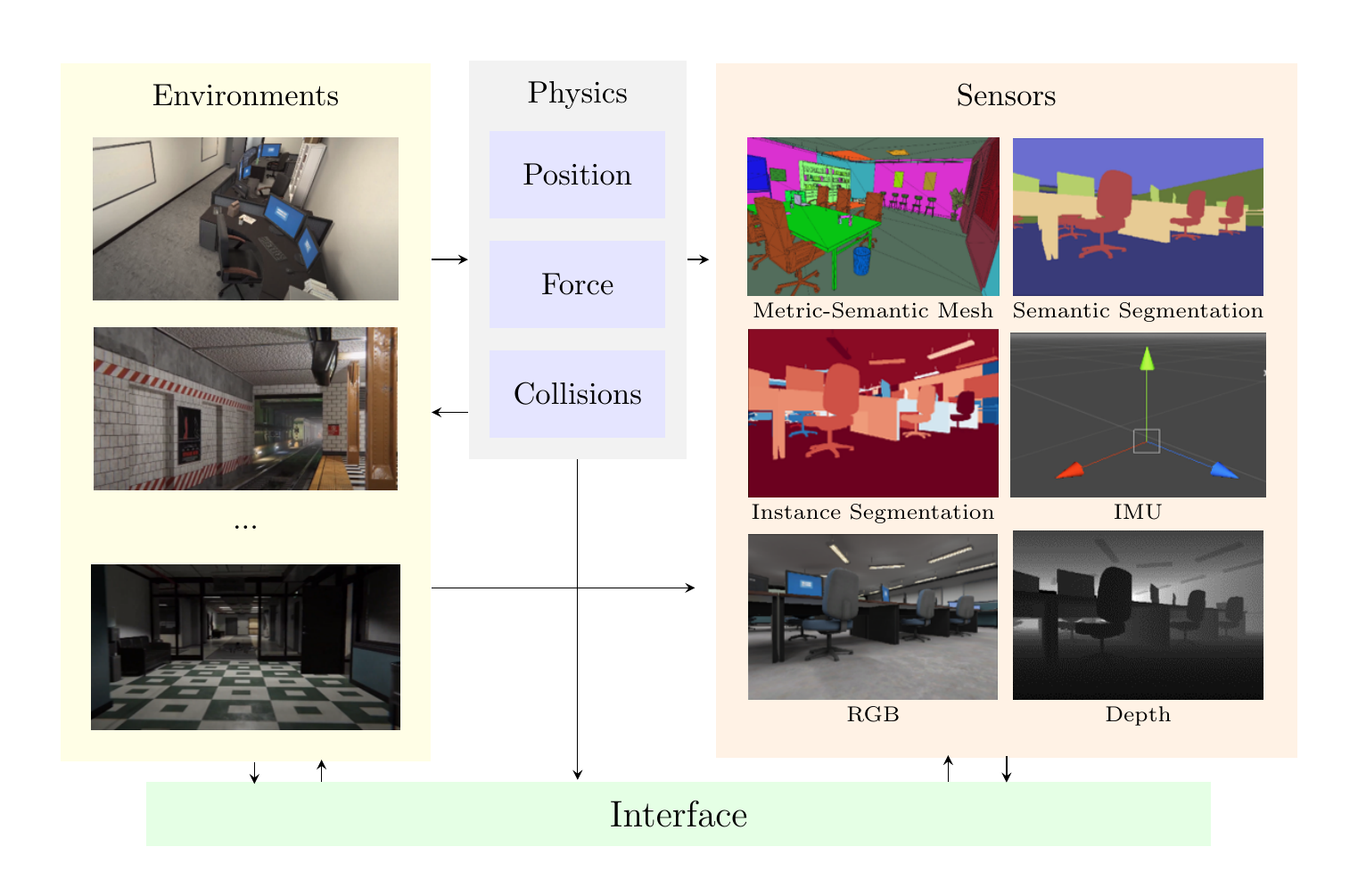}}
    \caption{TESSE is composed of four modular pieces. 1) Environments are built with 3D assets and are easily configurable. 2) The physics backend controls agent position in the environment, applies force, and detects collisions. 3) A variety of sensors provide rich 2D and 3D metric-semantic information. 4) An interface is used to change scenes, control the agent, and read sensory data. The Python interface can be integrated into frameworks such as ROS~\cite{ros} and OpenAI Gym~\cite{openai_gym}.}
    \label{fig:sim_overview}
\end{figure*}

\section{Related Work}
\label{sec:related_work}

\textbf{Simulation for Autonomous Systems.} Simulation accelerates autonomous system development where real world deployment is often costly, time consuming, and dangerous. Dosovitskiy et al.~\cite{pmlr-v78-dosovitskiy17a} created an autonomous driving simulator using 3D models and an underlying physics engines with exposed sensor streams. They compare several autonomous driving approaches including imitation learning, RL, and modular control over several navigation tasks. Several works incorporate vehicle dynamics models, such as FlightGoggles~\cite{8968116} and AirSim~\cite{shah2018airsim}, which provide a platform for developing UAV control and navigation algorithms. 

\textbf{Simulation for Computer Vision.} Simulation environments offer 3D structure and interactivity useful for many computer vision tasks. The Matterport3D \cite{Matterport3D} environment uses a large set of RGB-D panoramic views to create environments for a variety of tasks including keypoint matching and surface normal estimation. The 2D-3D-S \cite{armeni2017joint} dataset provides environments reconstructed from indoor scans for large scale scene understanding. Along with geometric information, an automated or semi-automated process is used to add semantic annotations. Armeni et al.~\cite{armeni_cvpr16} introduce the concept of semantic scene parsing. Given point clouds from indoor buildings, their method estimates several attributes including semantics, object bounding boxes, and room topology. The recently proposed scene graph paradigm captures hierarchical object relationships using an environment derived from registered panoramic images and external localization estimates to build a notion of global space~\cite{armeni20193d}.

\textbf{Simulation for Reinforcement Learning.} The reinforcement learning community has heavily leveraged simulation which, compared to the real-world, provides a scalable and safe method for training RL agents. One method for creating simulators is to design environments comprised of photorealistic 3D models as used by game engines. The AI2-THOR framework takes this approach to support tasks such as target driven visual navigation~\cite{ai2thor, zhu2017icra}. Alternatively, environments reconstructed from scanned panoramic images increase realism. Simulators such as Habitat~\cite{savva2019habitat} and Gibson~\cite{xia2018gibson} use this method to train agents for a variety of navigation tasks. Both modeled and reconstructed environments have incorporated interactivity, providing platforms for Interactive Question Answering (IQA) and Interactive Navigation~\cite{Gordon2017IQAVQ, xia2020interactive}.

While simulators have become increasingly sophisticated in terms of realism, sensor models, and vehicle dynamics, some applications combining perception and task execution leave capabilities to be desired. Existing simulation frameworks lack the combination of metric-semantic ground truth, continuous movement models, and photorealism while retaining lightweight interfaces and flexibility for use in multiple application areas. 

\section{Simulator}
\label{sec:simulator}

\begin{figure*}[t!]
    \centering
    \subfigure[Metric-Semantic Mapping]{
        \includegraphics[width=0.3\textwidth]{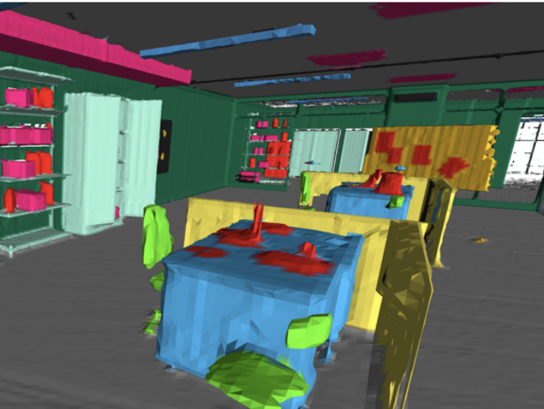}
    }  \hfill
    \subfigure[3D Dynamic Scene Graph]{\includegraphics[width=0.31\textwidth]{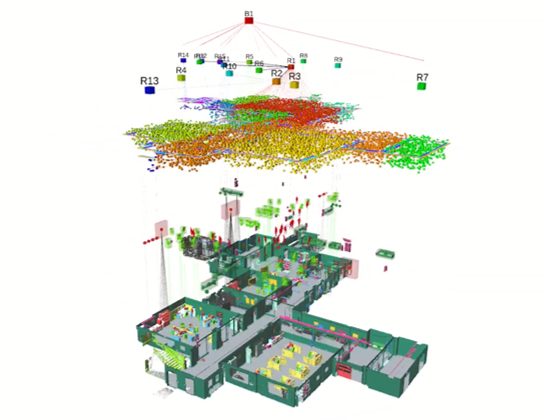}} \hfill
    \subfigure[Autonomous Racing]{\includegraphics[width=0.31\textwidth]{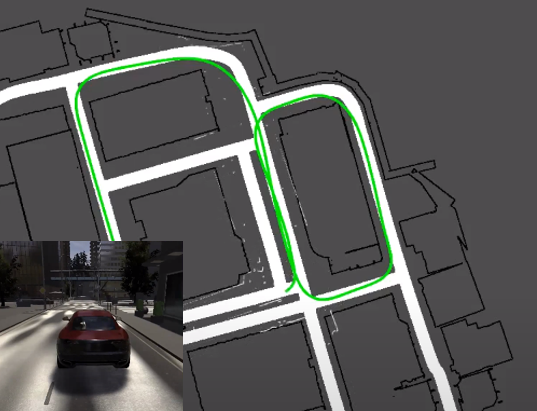}}
    \caption{Perception and task execution applications developed in TESSE: (a) Metric-semantic mapping creates a 3D, semantically-labeled map of the environment using only stereo cameras and an IMU \cite{Rosinol20icra-Kimera}. (b) 3D dynamic scene graph generation builds upon the metric-semantic map to find hierarchical relationships in an environment \cite{rosinol20203d}. (c) Autonomous racing requires control of a vehicle governed by a dynamics model using input from noisy sensor models.}
    \label{fig:tesse-applications}
\end{figure*}

TESSE simulates complex environments and provides a range of 2D, 3D, and inertial data products useful for developing perception and task execution algorithms. Built with Unity3D and designed in a modular fashion, TESSE can be quickly integrated into a variety of applications and is easily extendable. Illustrated in Figure~\ref{fig:sim_overview}, the simulator is composed of four major components. The environment contains the physical layout of a scene, a physics backend governs the movement of an agent, and a suite of sensors provide an agent with information about its environment. These pieces are exposed via an interface implemented in Python, making it compatible with common robotics and RL tools such as ROS~\cite{ros} and OpenAI Gym~\cite{openai_gym}. 

\textbf{Environments} are composed of 3D assets which can be either custom designed or imported. Typically, environment composition starts with a base layer that defines a scene layout. For example, in an indoor office this would consist of floors, walls, ceilings, etc. Then, environment specific objects such as tables, desks, and chairs are placed appropriately. Finally, details such as monitors, keyboards, and laptops may be added to provide further realism. TESSE also supports dynamic assets that, enabled by animation logic, realistically move around an environment 

Each asset contains geometric and semantic information. Thus, object and scene level data including 3D mesh, bounding boxes, and semantic labels are exposed. Objects are assigned varying semantic categories depending on the task. For example, obstacle avoidance may require course labels (i.e., obstacle or non-obstacle), whereas indoor target navigation could benefit from more detailed, object-level class information. These semantic labels may be changed easily from an external configuration.

\textbf{The physics backend} controls agent movement via force and torque commands which create continuous trajectories and generate inertial data. Discrete translations are also available to support higher level planning tasks. 

TESSE's frames per second (FPS) varies depending on device resources. Though inconsequential for many applications, this variation could change the behavior of frequency dependant processes such as control loops. Thus, TESSE may be run in \textit{step mode} which decouples the physics update rate from FPS, taking full advantage of available resources while keeping frequency dependent processes constant across devices.

\textbf{Sensors} provide 2D, 3D and inertial data as illustrated in Figure~\ref{fig:sim_overview}. 2D sensors include stereo, depth, semantic segmentation, and instance segmentation. 3D semantically annotated mesh is available on a per-scene basis. Due to the size of this data type, it is exported offline in a common format, such as a PLY or OBJ file, containing instance or semantically annotated vertex and edge data. A common coordinate system between the mesh and agent allows for the data to be processed interactively or offline. Odometry provides the agent's position, velocity, and acceleration in 3D space as well as collision information. To support high frequency applications, odometry is published over a high rate (200\,hz) broadcast.

\section{Applications}
\label{sec:applications}

\begin{figure*}[t!]
    \vspace{0.1in}  
    \centering
    \centerline{\includegraphics[trim=0 15 0 30,clip,width=1\textwidth]{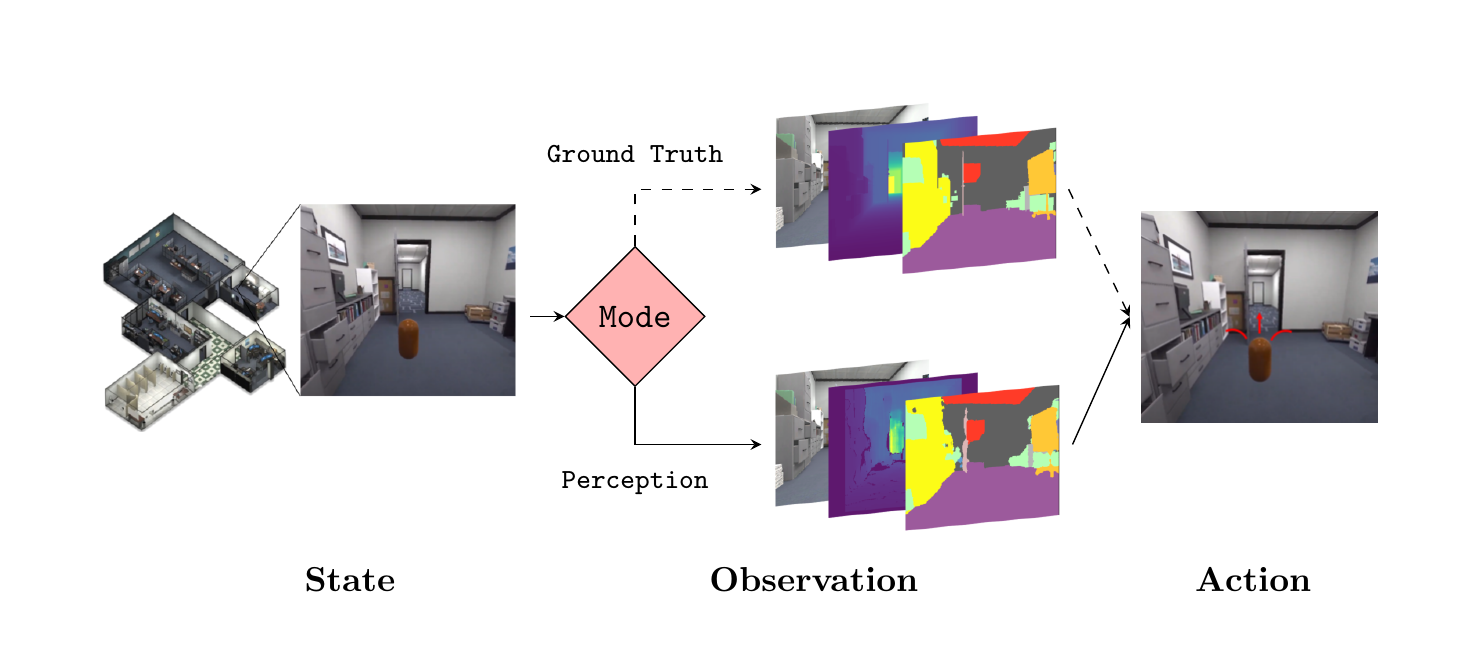}}
    \caption{In the GOSEEK Challenge, an agent must find targets placed within an office building. 1) State consists of its position and target locations. 2) RGB, depth, segmentation, and pose are provided as observations. In the Ground Truth track, these are read directly from the simulator. In the Perception track, these are estimated via a vision pipeline. 3) The agent must then choose one of four actions: forward, left, right, or attempt to collect a target.}
    \label{fig:goseek-overview}
\end{figure*}

TESSE has been used to develop perception algorithms including metric-semantic mapping and 3D dynamic scene graph generation~\cite{Rosinol20icra-Kimera, rosinol20203d}. TESSE was also used for the GOSEEK Challenge at ICRA 2020, a competition where participants developed agents to search for objects placed within indoor environments. 

\subsection{Metric-Semantic Mapping}
\label{sec:metric_semantic_mapping}

Metric-semantic mapping estimates the semantically labeled, 3D geometry of an environment. Such a representation may benefit several autonomy tasks including planning, navigation, and obstacle avoidance. Creating a fully autonomous metric-semantic mapping system presents several challenges. If depth is not directly provided by a sensor such as an RGB-D camera, it must be estimated to infer the 3D structure of a scene. Semantic labels in 3D space must be either directly estimated or projected from a 2D image. Local metric-semantic estimates must then be fused over a trajectory to provide a global map. Previous approaches use static datasets that provide the 3D structure of a scene to evaluate the SLAM component~\cite{oleynikova2017voxblox}. If a semantic component is included, portions of the dataset are manually labeled for evaluation~\cite{semantic_fusion, mask_fusion, segmap2018}. This approach is inherently limited and subject to noise. In contrast, TESSE provides the ability to create trajectories with data streams to create a metric-semantic map. The availability of perfect 3D and semantic ground truth simplifies evaluation. 

Rosinol et al.~\cite{Rosinol20icra-Kimera} use TESSE to develop a metric-semantic mapping approach. They estimate scene geometry via stereo reconstruction and bundled raycasting.VIO incorporates stereo and IMU data to estimate pose along a trajectory. This pose is used to localize the agent and allows for global mesh estimation from local estimates. 2D pixel-wise semantic labels are then projected onto the scene, producing a mesh such as in Figure~\ref{fig:tesse-applications}. Available ground truth semantic mesh enabled precise and accurate evaluation. Ablation studies were performed to evaluate each component in the system. During evaluation, ground truth pose was substituted for VIO and ground truth depth for dense stereo. 

\subsection{3D Dynamic Scene Graph Generation} 

3D dynamic scene graphs provide hierarchical relationships between entities in an environment, producing abstractions useful for high-level task planning. Ideally, such representations would allow for both high-level (e.g. ``find the remote in the living room") and low-level tasks (e.g., obstacle avoidance). Existing approaches use a semi-automated labeling pipeline and rely on registered images in a static environment, preventing fully autonomous execution ~\cite{armeni20193d}. 

Rosinol et al.~\cite{rosinol20203d} use the TESSE environment to create the first automatic 3D dynamic scene graph generator, the Spatial PerceptIon eNgine (SPIN), illustrated in Figure~\ref{fig:tesse-applications}. The graph consists of five layers: metric-semantic mesh, objects and agents, places and structures, rooms, and buildings. This method leverages several unique components of TESSE. The graph is constructed using data obtained from a moving agent with stereo and IMU sensors. The method described in Section~\ref{sec:metric_semantic_mapping} is used to create a metric-semantic mesh, forming the backbone of the scene graph. The constructed mesh is compared to a catalogue of CAD models provided by the simulator and used to estimate object bounding boxes and pose. Walking human agents are added to the environment and are tracked over time.

\subsection{Autonomous Driving}

Autonomous driving is a difficult, heavily researched robotics problem requiring perception, planning, and control. Solutions must reason about data from noisy sensors and account for uncertainly due to vehicles, pedestrians, or imperfect maps of the environment. Furthermore, fielding solutions on physical platforms can be slow and dangerous. Given these concerns, simulation is an ideal space to develop autonomous driving algorithms. 

TESSE provides a platform for autonomous driving research and was recently used to host an autonomous racing competition.\footnote{ http://news.mit.edu/2020/photorealistic-simulator-made-mit-robot-racing-competition-live-online-experience-0609} Using data from a set of sensor models and predefined road map, participants created a navigation solution to control a vehicle governed by a realistic dynamics model. A LIDAR sensor model provided depth estimates on the horizontal plane. The competition was held in the Windridge City environment~\cite{shah2018airsim} and featured two tracks. First, participants were tasked with completing the track in minimal time. Next, dynamic obstacles were added to the scene and participants had to complete the track with minimal collisions. The platform supported over 100 submissions on low-resource platforms. 

\subsection{GOSEEK Challenge}
\label{sec:challenge_overview}

The GOSEEK Challenge, hosted in conjunction with the Perception, Action, Learning Workshop at the International Conference of Robotics and Automation (ICRA) 2020,\footnote{ https://mit-spark.github.io/PAL-ICRA2020/} combined perception and high level decision making for object search in complex indoor spaces. Participants developed agents to search for targets randomly placed within TESSE environments, an example of which is illustrated in Figure~\ref{fig:goseek-overview}. Several ground truth data modalities were provided from the simulator for development and, in the Ground Truth track, used for evaluation. However, when operating on a physical platform perfect data cannot be expected. Thus, to motivate generalization capabilities, realistic corollaries for each modality were provided in a second mode coined the Perception track. The contest was hosted on the EvalAI Platform~\cite{DBLP:journals/corr/abs-1902-03570} where participants submitted solutions for on demand scoring. 

\textbf{Problem Definition.} The agent was spawned in an environment and had to find randomly placed targets within the allotted episode length. Environments were modeled after office floors and contained hallways, storage rooms, conference rooms, offices, and bathrooms. Targets were placed only within offices. Thus, to make the best use of time, the agent had to distinguish between offices and other types of rooms. The episode ended after either all targets were found or the agent took 400 steps, roughly the number of steps required by a human expert to collect all the targets.

A successful agent had to explore previously unseen environments which required some notion of memory to avoid revisits. Because targets were only placed in certain types of rooms, the agent had to pick up on semantic cues so as to ignore dead ends. Then, having visited the correct room, it had to identify and navigate to targets while avoiding obstacles. An episode limit required that all this be done in an efficient manner. 

\textbf{Observations.} RGB images, pixel-wise depth, pixel-wise semantic segmentation, and pose in 2D space were available as observations to the agent. Segmentation classes were defined as $\mathcal{C} \in \{$ \verb|floor|, \verb|ceiling|, \verb|wall|, \verb|monitor|, \verb|door|, \verb|table|, \verb|chair|, \verb|storage|, \verb|couch|, \verb|clutter|, \verb|target| $\}$. Ground truth data from TESSE was available for training and submission to the ground truth track. When running in the Perception track, the above modalities were replaced with estimates from the perception pipeline discussed in Section \ref{sec:multimodal_perception} below.

\textbf{Available Actions.} The agent selected from one of four actions at each step: \verb|move_forward|, \verb|turn_left|, \verb|turn_right|, and \verb|collect|. These action corresponded to the agent moving forward 0.5\,m, turning left 8 degrees, turning right 8 degrees, and attempting to collect the target. A target collect was considered successful if the target was within 2\,m of the agent and in its field of view. 

\textbf{Scoring.} Training in simulation presents a variety of challenges. An agent may learn behavior that exploits unrealistic simulation features which can be detrimental to deployment on a physical system~\cite{Kadian2019AreWM}. For instance, in the real world, beyond simply finding the maximum number of targets, we may be concerned with path efficiency, precision, and avoiding collisions. For this reason, we included several factors in the score defined as

\begin{equation*}
 s = r + w_p p - w_c \frac{c}{l} - w_a \frac{a}{l}\,,
\end{equation*}

where $r$ is target recall, $p$ is precision of attempted collects, $l$ is episode length, $c$ is the number of collisions in an episode, and $a$ is the number of actions taken. $w_p$, $w_c$, and $w_a$ are weighting factors. We used $0.1$ for all weighting factors in this competition. 

\textbf{Evaluation.}  We used Monte Carlo evaluation over 100 episodes on two scenes withheld for testing to estimate an average episode score. Participants developed solutions in Docker containers which, once completed, were uploaded to the EvalAI platform~\cite{DBLP:journals/corr/abs-1902-03570} and run on Amazon Web Services. 

\subsubsection{Multimodal Perception}
\label{sec:multimodal_perception}

To motivate robust and generalizable solutions, the Perception track provided depth, segmentation, and pose estimates from computer vision models in lieu of ground truth. The noisy perception estimates were intended to mimic the quality of observations available to an agent when deployed in the real world. 

\textbf{Visual Intertial Odometry} with Kimera-VIO~\cite{Rosinol20icra-Kimera} provided pose estimates given stereo and IMU data from TESSE. Although the agent planned over discrete actions, VIO required a continuous trajectory. Thus, discrete steps were translated to continuous movement via a proportional-derivative controller. This added a further element of realism since discrete movement commands were not be perfectly executed. 

\textbf{Depth} was estimated via stereo reconstruction. Stereo block matching was used to produce a disparity image point cloud. Points were then projected into the left camera plane to produce a depth image aligned with the provided ground truth.
Stereo reconstruction introduced several sources of noise. 

\textbf{Segmentation} was provided by a U-Net~\cite{RFB15a} model trained with images from the first four GOSEEK scenes. The fifth scene was used to collect a test set. Due to the plentiful data available in simulation, the network could achieve near perfect accuracy. However, to provide realism, training was stopped early to achieve a mean Intersection-over-Union (mIoU) score of 0.81. The model was then exported to TensorRT for real-time inference.

\subsubsection{Baseline RL Policy}

\begin{figure}[t]
    \vspace{0.1in}  
    \centering
    \includegraphics[width=1\textwidth]{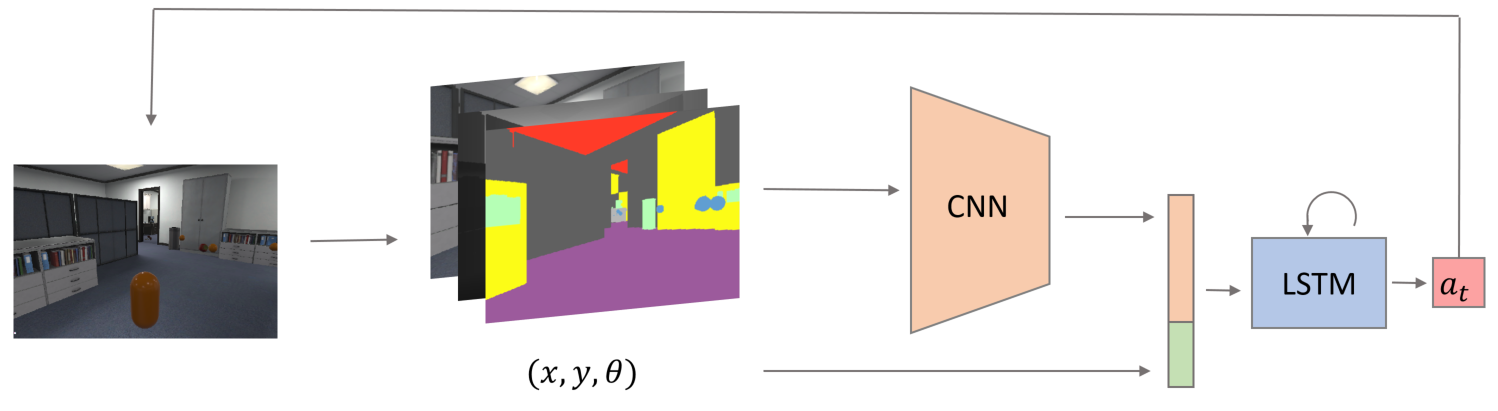}
    \caption{GOSEEK baseline policy network. A CNN extracts image features. The pose vector is passed through a dense layer. The CNN and pose features are concatenated and used by an LSTM to predict an action.}
    \label{fig:gosee-baseline}
    \vspace{-0.1in}
\end{figure}

We provided a baseline model trained with Proximal Policy Optimization (PPO)~\cite{DBLP:journals/corr/SchulmanWDRK17}. The policy network, illustrated in Figure~\ref{fig:gosee-baseline}, took as observation RGB, depth, and segmentation images along with pose. The images were stacked and fed into a Convolutional Neural Network (CNN) with three layers of size 32, 64, 64, followed by a dense layer of size 512. The output was flattened and concatenated with the pose vector. The resulting feature vector was then given to a Long Short-Term Memory Network (LSTM) with 256 hidden units which output a distribution over available actions. We then sampled from this distribution to determine the agent's next step.

The agent was given a reward of $+1$ for each target found. To encourage short trajectories, a small time penalty of $-0.1$ was given at every step. The policy was trained for $5e6$ steps using the Adam optimizer~\cite{adam}. A validation scene was used to select the best policy at test time. 

\subsubsection{Results}
Table \ref{table:goseek-results} reports the results of our baseline policy in the Ground Truth and Perception tracks. For comparison, we report the performance of a human participant in the Ground Truth track. To provide a lower bound on performance, we also report the result of a random policy that takes each available action with an equal probability. Evaluation was performed in two holdout scenes over 100 episodes. 

\nopagebreak[2]

\begin{table}[h]    
    \vspace{0.1in}
    \setlength{\tabcolsep}{3pt}
    \renewcommand\cellalign{lc}
    \newcommand{\hackspacing}{\\ \vspace{-7.5mm} \\}  
    \centering
    \begin{tabular}{lrrrrr} 
     \toprule
     Approach & Recall & Precision & Collisions & Steps & Score \\
     \midrule
     \makecell{Baseline in \\Ground Truth} & 43.4\% & 21.3\% & 76.4 & 400 & 0.34\\
     \hackspacing
     \makecell{Baseline in \\Perception} & 30.5\% & 18.7\% & 200.5 & 400 & 0.17\\
     \midrule
     Human & 88.9\% & 95.8 \% & 11.7 & 385.9 & 0.89 \\
     \hackspacing
     Random & 5\% & 1\% & 256.2 & 400 & -0.12\\
     \bottomrule
    \end{tabular}
    \vspace{1mm}
    \caption{GOSEEK Average per-episode results.}
    \label{table:goseek-results}
\end{table}

\begin{figure}[h!]
    \centering
    \newcommand{\trajsize}{0.47}
    \newcommand{\trajspace}{-0.02}
    \subfigure[Ground Truth]{\includegraphics[width=\trajsize\textwidth]{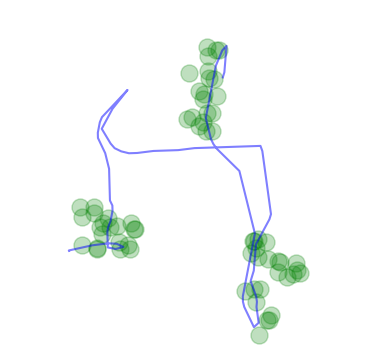} }
    \hspace{\trajspace\textwidth}
    \subfigure[Perception]{\includegraphics[width=0.45\textwidth]{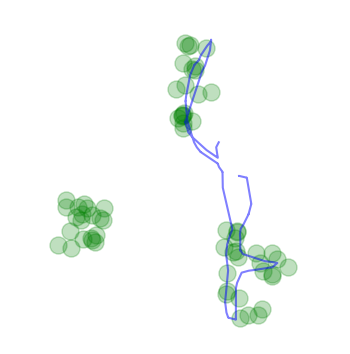}} %
    \\
    \hspace{\trajspace\textwidth}
    \vspace{0.2in}
    \subfigure[Human]{\includegraphics[width=\trajsize\textwidth]{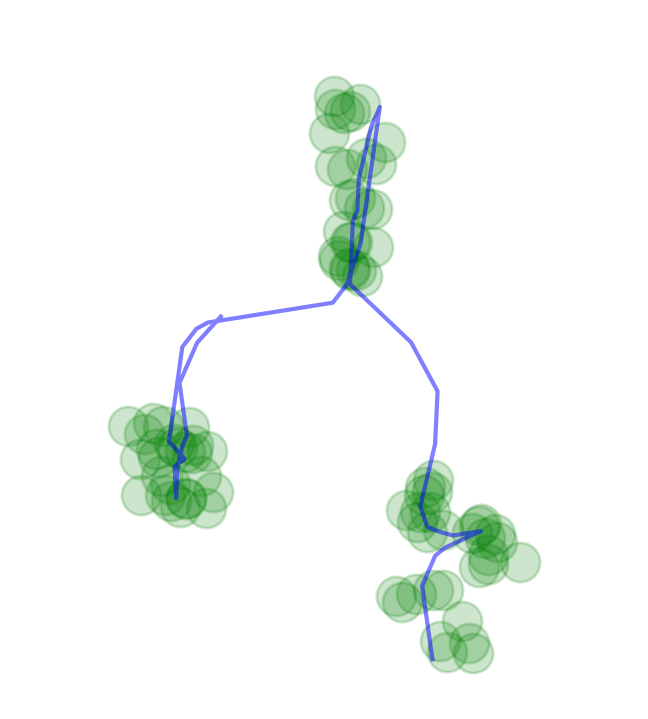}}
    \hspace{\trajspace\textwidth}
    \subfigure[Random]{\includegraphics[width=\trajsize\textwidth]{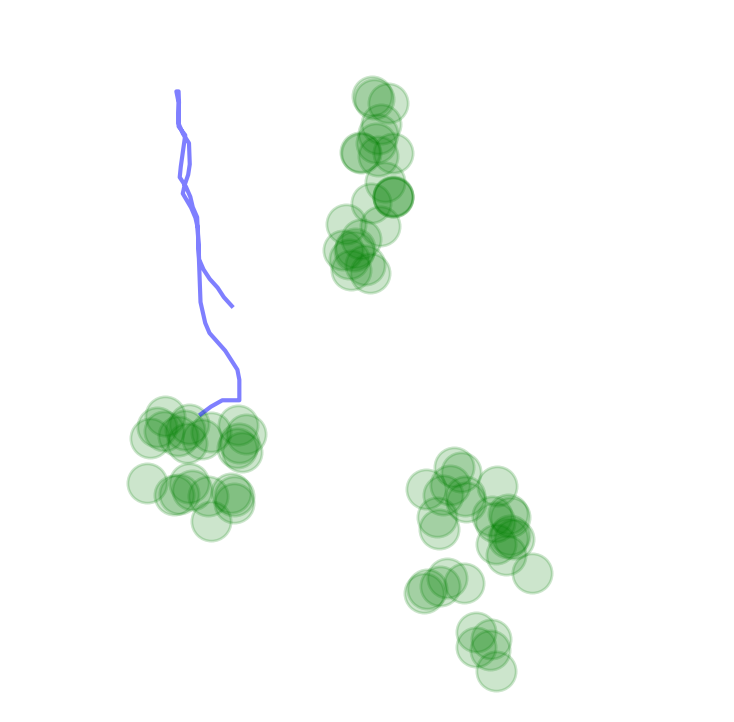}}
    \caption{Example trajectories, target distributions are indicated in green. The baseline policy explores more area in the Ground Truth track as compared to the Perception track. In the latter case, increased collisions and imperfect actuation hinder navigation.}
    \label{fig:trajectories}
\end{figure}

Unsurprisingly, the baseline policy performed best on the Ground Truth track. There is minimal distribution shift between the training and testing domains and actuation is perfectly realized. In many cases, the policy exhibited desirable performance by successfully entering offices, navigating to targets, and collecting them.

Nevertheless, the baseline policy was far from perfect. When no targets were in the policy's vicinity, it often had difficulty exploring new space in the environment. Thus, some episodes ended with unvisited offices. The policy frequently doubled back on its trajectory and revisited rooms, indicating that it did not have a robust sense of memory. Furthermore, despite access to depth information, the policy often attempted to collect targets out of range which led to low precision.

Noisy observations introduced in the Perception track lead to a significant decrease in recall. Precision did not drastically decrease, indicating that even when using noisy data the baseline policy could recognize targets in close proximity. More significantly, the policy could not effectively explore the scene with noisy perception. Policies in the Perception track visited significantly less space, as indicated in Table \ref{table:explored_space} and illustrated in Figure~\ref{fig:trajectories}. Average collisions more than doubled, suggesting that the policy's ability to navigate in tight spaces also degraded. Imperfect actuation introduced from the continuous controller likely exacerbated this problem. 

\begin{table}
    \vspace{0.1in}
    \centering
    \begin{tabular}{lrrrrr} 
     \toprule
     Approach & Explored Space \\
     \midrule
     Baseline in Ground Truth & $54.3 \pm 16.1$ m$^2$ \\
     Baseline in Perception  & $42.5 \pm 13.8$ m$^2$\\
     \midrule
     Human & $64.3 \pm 14.0$ m$^2$ \\
     Random & $11.8 \pm 7.2$ m$^2$\\
     \bottomrule
    \end{tabular}
    \vspace{1mm}
    \caption{GOSEEK average per-episode space explored.}
    \label{table:explored_space}
\end{table}

\section{Conclusion}
We present TESSE, a simulator for bridging spatial perception and task execution. Through interactively providing 2D, 3D, and inertial data, TESSE enables the development and evaluation of both complex perception systems and task execution approaches. We demonstrate TESSE's utility through metric-semantic mapping~\cite{Rosinol20icra-Kimera}, 3D dynamic scene graph generation~\cite{rosinol20203d}, autonomous racing, and the GOSEEK challenge at ICRA 2020, all of which used this simulation framework. By open sourcing the simulator and associated examples, we hope to aid research in these areas. 

To date, TESSE has supported scene understanding algorithms that produce hierarchical, 3D representations of the environments. Concurrently, autonomous racing and the GOSEEK challenge incorporate 2.5D data sources into control and navigation tasks. We believe that future work combining 3D data products into task execution pipelines presents an exciting opportunity to create robust and generalizable agents.


\section*{ACKNOWLEDGMENT}

The authors would like to thank the members of the MIT Lincoln Laboratory Technology Office
and Autonomous Systems Line Committee for their support. We also thank Mark Mazumder for his valuable feedback on our manuscript.


\bibliography{references}
\bibliographystyle{plain}

\end{document}